# Improved Reinforcement Learning through Imitation Learning Pretraining Towards Image-based Autonomous Driving

Tianqi Wang[1] and Dong Eui Chang[1*]

[1] School of Electrical Engineering, Korea Advanced Institute of Science and Technology,
Daejeon, 34141, Korea (tianqi_wang@kaist.ac.kr, dechang@kaist.ac.kr)* Corresponding author

**Abstract**: We present a training pipeline for the autonomous driving task given the current camera image and vehicle speed as the input to produce the throttle, brake, and steering control output. The simulator Airsim's [1] convenient weather and lighting API provides a sufficient diversity during training which can be very helpful to increase the trained policy's robustness. In order to not limit the possible policy's performance, we use a continuous and deterministic control policy setting. We utilize ResNet-34 [2] as our actor and critic networks with some slight changes in the fully connected layers. Considering human's mastery of this task and the high-complexity nature of this task, we first use imitation learning to mimic the given human policy and then leverage the trained policy and its weights to the reinforcement learning phase for which we use DDPG [3]. This combination shows a considerable performance boost comparing to both pure imitation learning and pure DDPG for the autonomous driving task.

## 1. INTRODUCTION

For autonomous driving or navigation, the problem of generating control decision given a monocular camera image and other auxiliary vehicle sensors' outputs such as vehicle velocity, acceleration etc. has aroused a lot of interest in recent years [4, 5, 6, 7, 8, 9]. Due to the huge development of deep learning and reinforcement learning in recent years, most of the recent works towards this task involve these two methodologies.

Dealing with the high-dimensional image input, CNN(Convolutional Neural Network) is commonly used to act as a feature extractor to generate useful features for subsequent layers. There are many well-accepted CNN networks for robotics usage, such as ResNet [2], VGG-Net [10], R-CNN [11], etc. More importantly, these CNN models have been proved to be able not only to solve a specific task, but also to handle other tasks thanks to their feature extractor nature.

Driving through input camera images is a task at which humans are relatively good. As a result, conducting imitation learning towards given human policy might produce a relative well-performed policy. However, due to the limitation on the possible policy performance and potential lack of generalizability for unseen situations, not many researches have focused on leveraging this method to this task. Instead, using reinforcement learning (RL) in a simulator and then transfer the learned policy to the real world is very popular in recent research [4, 7, 8, 9]. Reinforcement learning has no limit on the possible learned policy's performance, and its performance heavily depends on the reward function design and the exploration coverage of the possible state and action spaces. Nonetheless, when the state and action spaces are huge or high-dimensional, pure RL algorithm might never get a sufficient amount of useful interaction experiences and the training process usually diverges. Surprisingly, not many researches focus on combining these two methods and thus leverage both their advantages for this image-based autonomous driving task.

## 2. RELATED WORK

### 2.1 Image-based autonomous driving or navigation

This kind of tasks involving the utilization of deep neural networks are usually trained in an end-to-end style. Some research implements supervised learning (or sometimes called imitation learning) using annotated labels [5]. Others use reinforcement learning [4, 7, 8, 9] through interacting with the environment and use the sampled reward as the supervision signal to indicate how good or bad is a specific action at a specific state.

For the supervised/imitation learning methods, they struggle with the difficulty to collect sufficient data with enough diversity for some tasks and can only perform at most as well as the supervisor. For the data diversity issue, there are many possible solutions. First is the well-known data augmentation process, such as random cropping, drifting, lighting change, etc. Second is to randomize the environment during training such as weather and lighting conditions, object types, positions and texture as shown in [4]. In [5], the author uses three cameras installed on the left, middle and right side of a drone and keeps the drone facing forward open space all the time, then labels the images collected by left, middle, right cameras as 'go right', 'go straight', 'go left' respectively. However, this extremely simple control policy setting directly constraints the possible performance.

As for reinforcement learning, the randomization of the training environment also affects the robustness and performance of the possible trained policy heavily. Besides that, proper reward function design is crucial for the agent to extract useful information from its past interaction experiences with the environment. For example, some tasks with only sparse reward functions (0 or 1) are still very challenging or even unsolvable. For this autonomous driving task, knowing the distance to the nearest obstacle is the most important component of the reward function as mentioned in [4].

### 2.2 CNN networks for robotic vision

Due to the success of recent CNN networks in traditional computer vision tasks like classification or object detection, more and more researchers implement these existing networks into robotic vision task and have achieved considerable performance. Autonomous driving based on the camera image input is clearly one of them.

Besides those novel and complete architectures we mentioned above, such as ResNet [2], VGGNet [10], R-CNN [11], etc., some work also focuses on the design of integrable modules to extract more useful information from the existing architecture, such as the attention module [12, 13, 14]. The attention module can be classified into the channel (RGB) and spatial (pixel coordinates) attention parts, which emphasize more on specific channels or spatial regions. Since human's decision during driving clearly comes more from some specific parts of the input image, the integration of these attention modules might produce a better performance.

### 2.3 Reinforcement learning for deterministic and continuous policy

Considering the possible catastrophic effect for undesired behavior in this task, clearly deterministic policy setting is a better choice in comparison to the stochastic counterpart. Besides that, unlike the discretized possible actions in [4, 5, 8, 9] which can be very likely to limit the possible performance, ideally, continuous policy setting is favorable for this task.

As for the deterministic and continuous policy-based reinforcement learning algorithms, DDPG [3] is the most widely used algorithm and has shown impressive results in many less complex tasks. It, however, can cost too much training time or never converge in complex problems. Given the high-dimensional nature of the state input, we expect that pure DDPG might not work well in this task. As a result, a combination with imitation learning becomes more important.

### 2.4 Two ways to combine imitation learning with reinforcement learning

The most natural and direct idea is pre-training and fine-tuning, which uses the pre-trained weights as the initial weights and use reinforcement learning to fine-tune them. The other way is to use a hybrid objective during reinforcement learning, in which one of the components is related to imitation learning. For efficiency and simplicity, we choose the first method in this paper.

## 3. PROPOSED METHOD

Humans are actually relatively good at driving a car simply based on the current given image whether in real life or in racing games. Moreover, human's driving policy are quite robust with the changing weather or lighting conditions. Thus, imitation learning with a human's driving policy as a supervision signal can generate a relatively good initial policy for the subsequent reinforcement learning phase, which can significantly reduce the training time, stabilize the training process and produce better results than training from scratch using pure reinforcement learning for this kind of high-dimensional input problems.

For this autonomous driving problem, deterministic policy setting is clearly a better choice than the stochastical policy setting since this is a problem where undesired behavior might produce a catastrophic consequence almost immediately such as collision. Besides that, unlike other papers [4, 5, 8, 9] which discretize the possible action space and then use value-based reinforcement learning algorithms which enjoy the reduction of training difficulty with the very likely sacrifice in the performance as well, we choose the continuous action space setting and implement a policy-based RL algorithm.

### 3.1 Imitation learning

To generate a good initial policy from imitation learning, we have to make sure that the dataset used for supervising must have tackled with situations with enough diversity such as going straight and accelerating when the forward space is open and steering to avoid the collision or going through the corners under different weather and lighting conditions. How much the supervising dataset explores the environment would play a vital role in the produced policy's performance and generalizability.

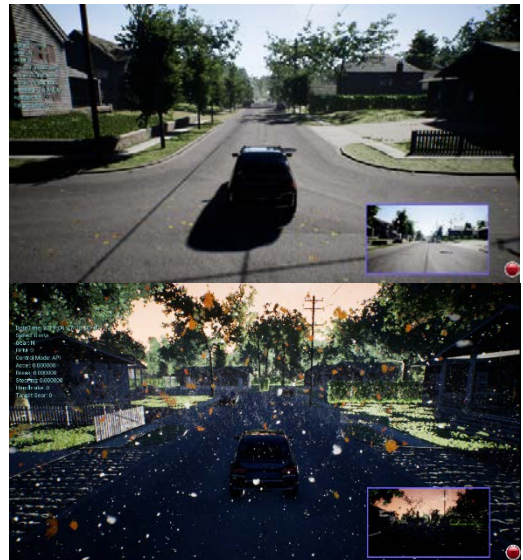

Fig. 1 Various weather and lighting conditions in Airsim [1]

We use an Xbox game controller to control the car and record the data using Airsim's API in the background. To ensure the dataset quantity balance, we predefine planned sample numbers under different weather and lighting conditions evenly and try to experience enough situations with similar respective sample numbers in different weather and lighting conditions.

Considering the factors that affect human's decision during driving, apart from the current image input, people would also try to get a roughly estimated vehicle speed based on the recent consequent input images to guide the decision making. Based on that, we consider current image input and current vehicle speed as our current state inputs, and throttle, brake from zero to one and steering from minus one to positive one as our control outputs.

For the loss used in the imitation learning phase, we choose the Huber loss in Eq. (1) which can be seen as a combination of L1 and L2 loss, is less sensitive to outliers and can make the training process more stable compared to MSE(Mean Squared Error) as a result.

$$Loss(y,\hat{y}) = \begin{cases} 0.5 * (y - \hat{y})^2, & |y - \hat{y}| \leq \delta \\ \delta(|y - \hat{y}| - 0.5 * \delta), & otherwise \end{cases} \quad (1)$$

### 3.2 Network architecture

The success for deep learning in recent years comes largely from the development of deep neural networks especially the CNN-based architectures which are usually seen as feature extractors for the high dimensional input and has been proved to be useful not only for a specific task; refer to [15] for CNNs.

As Fig. 2 shows, here we choose ResNet-34 [2] as the backbone structure for both actor and critic networks with slight differences in the fully connected layers because of the input dimension difference. After ResNet-34's average pooling layer, we choose to have three fully connected layers for both actor and critic networks. The velocity input is concatenated with the output of ResNet-34's average pooling layer for both actor and critic networks which can be seen as the extracted features of the input image. The chosen action (throttle, brake, steering) is concatenated in the final fully connected layer's input such that the actor and critic networks have identical structures except in the last fully connected layer due to the involving action input in the critic network.

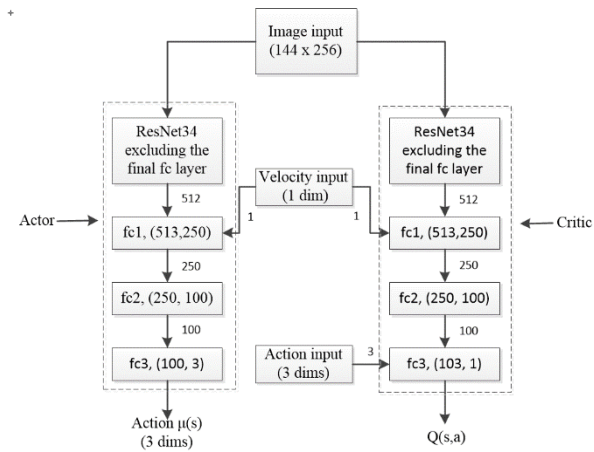

Fig. 2 Actor and critic networks

As a comparison, we also try ResNet-34 and ResNet-50 [2] with several attention modules, like SE (Squeeze-Excitation Module) [12], BAM(Bottleneck Attention Module) [13] and CBAM(Convolutional Block Attention Module) [14] which emphasize rather channel or/and spatial attention.

However, surprisingly, as shown in Table 1, simple and light ResNet-34 [2] outperforms other architectures in terms of both final testing loss and the generated policy's performance. Based on this, we choose ResNet-34 [2] as our architecture for actor and critic.

Table 1 Testing loss and generated policy's performance for different architectures.

| Architecture | Testing loss | Average accumulated reward |
|---|---|---|
| ResNet-34 [2] | **0.0835** | **79.82** |
| + SE [12] | 0.1211 | 50.14 |
| + BAM [13] | 0.1105 | 44.02 |
| +CBAM [14] | 0.1245 | 18.73 |
| ResNet-50 [2] | 0.1037 | 30.26 |
| + SE [12] | 0.1344 | 20.11 |
| + BAM [13] | 0.1356 | 8.78 |
| +CBAM [14] | 0.1471 | 5.12 |

### 3.3 Reinforcement learning for performance boost

The drawback of imitation learning is obvious because adequate expert guidance can sometimes be hard to acquire, or more importantly, the learned policy can only perform at most as well as the guiding expert's policy. On the other hand, reinforcement learning works based on using the exploration and exploitation interaction experience with the environment, thus has no limit on the possible performance. However, pure reinforcement learning can suffer from huge and high dimensional possible state and action space which can result in an insufficient exploration or prohibitively long training time.

Thus combining imitation learning and reinforcement learning can be an efficient and promising method for complex problems for which expert guidance are possible.

Instead of the true label used in supervised learning or imitation learning, we can only get reward as the supervision signal in reinforcement learning to evaluate how good is the previous action in the previous state. As a result, the choice of the reward function would heavily affect the possible performance of the learned agent. Here, we consider the distance to the nearest obstacle (the road edges are considered as obstacles as well) and current vehicle speed as two components of the reward function to make sure that only keeping far from the obstacles and driving fast at the same time can result in a high reward. The reward function we design is shown in Eq. (2).

$$r = \begin{cases} 0, & if\ r_{distance}\ or\ r_{speed} < 0.1 \\ \lambda_d * r_{distance} + \lambda_v * r_{speed}, & otherwise \end{cases}$$
$$r_{distance} = \max(1, d/d_\theta) \qquad , \quad (2)$$
$$r_{speed} = \max(1, v/v_\theta)$$

where $d_\theta$, $v_\theta$ are the threshold values we choose for the ideal distance from the obstacles and vehicle speed, $\lambda_d$, $\lambda_v$ are the weights of reward's distance and speed component which should have a sum of one. In this paper, we set $d_\theta = 3.5$ m, $v_\theta = 20$ m/s, $\lambda_d = \lambda_v = 0.5$. Based on this setting, our reward function is squeezed to the interval [0,1].

Based on our continuous and deterministic control policy setting, we choose DDPG (Deep Deterministic Policy Gradient) [3], one of the most popular RL algorithms for continuous and deterministic control policy setting, as our reinforcement learning algorithm.

However, in order to combine imitation learning with DDPG, we make several changes on the original DDPG algorithm:

- For the initial weights of the actor and critic networks, we give them the pre-trained weights except for the final fully connected layer of imitation learning's generated actor network. The actor network used in the imitation learning phase and that used in the reinforcement learning phase have the same architecture while the critic network's final fully connected layer is different from the actor network. Actually, we can give all the weights of the pre-trained actor network to the actor networks used in the DDPG and give all the weights except the final fully connected layer to the critic networks. However, we have found that since in the original pure DDPG, the actor and the critic are updated synchronously so that the interplay between these two can guide them to get better weights, giving the actor networks of DDPG the whole weights of the pre-trained actor network can result in an out-of-sync problem for the actor and critic, which would make the training process diverge. Besides this, we also freeze all the convolutional layers and only train the remaining fully connected layers to stabilize the training and shorten the training time.
- For the experience replay pool, we first fulfill it using the samples collected under imitation learning's generated policy and use this experience replay pool to first train the actor and critic networks until the value loss and policy loss converges. After that, we follow the normal DDPG to collect new experience from the updated actor network and update the actor and critic networks iteratively.
- The OU(Ornstein–Uhlenbeck) noise used in the original DDPG is aimed at introducing more exploration towards the state and action space especially in the early stage of the training process and the effect of the OU noise is designed to gradually decrease to zero as the training process progresses. In our case, the experience replay pool collected under imitation learning's generated policy already contains sufficient useful information about the good and bad experience with the environment. Thus, we consider removing the OU noise for this problem.

## 4. EXPERIMENTS

### 4.1 Imitation learning phase

For the benefits of using imitation learning as the preceding phase for RL besides what we mentioned before such as stabilizing the RL training process and shortening the training time, there is another very important advantage which is giving us the opportunity to check whether the chosen architecture is able to represent a reasonable policy (provided by human) or not.

Originally, we use an Xbox controller to control the car under various weather and lighting conditions and record the data using Airsim API. During recording, we drive the car in a continuous style which means we almost never stop and depart again. This causes an interesting problem which is that the generated policy always brakes hard and do not accelerate when facing corners even if the current vehicle speed is almost zero. This problem is quite understandable since we do not provide this kind of samples in the training dataset. As a very natural idea to augment the dataset, we set all the samples' current vehicle speed to a random number from 0 to 3 m/s and the throttle control output to 1, brake to 0 and steering angle unchanged to get augmented dataset.

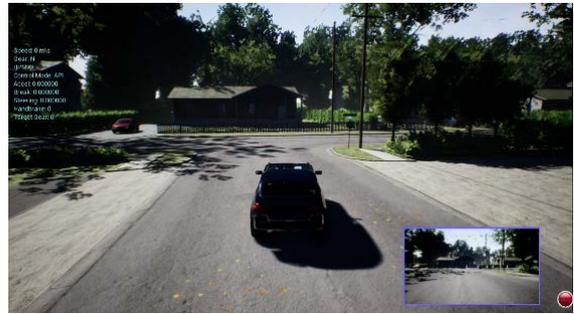

Fig. 3 Cornering situation

After this dataset augmentation, the generated policy always accelerates if the current vehicle speed is very low and remains all the good behavior that the previous trained policy under the original dataset has. As a result, this imitation learning phase shows that our chosen ResNet-34 architecture is able to take both the current image and vehicle speed into account to get a good control policy.

As the second picture of Fig. 1 shows, the input image during the night is quite unclear and we find that during testing none of the trained policy works well during the night. For simplicity, we only use the daytime setting with various weather conditions for testing and subsequent reinforcement learning phase. But still, in imitation learning, we use all the data recorded during both daytime and night to get a more robust policy.

For testing the trained policy's performance, we set the maximum number of timesteps to 1000 for each episode and terminates when a collision happens. We randomly choose the weather condition and the initial position of the car before each episode starts and calculate a 5 episodes' average accumulated reward as the criteria. As shown in Table 1, the simple and light ResNet-34 outperforms other architectures with attention module integrated in terms of both final testing loss and the generated policy's performance. Based on this, we choose ResNet-34 as our architecture for actor and critic.

## 4.2 Reinforcement learning using pre-trained weights

Similarly, during training and testing of the reinforcement learning phase, we set a maximum number of timesteps to 1000 for each episode and terminates when a collision happens. We randomly choose the weather condition and the initial position of the car before each episode starts and calculate 5 episodes' average accumulated reward as the criteria.

We compare the testing performance of the three policies: the original imitation learning's generated policy, the policy obtained from pure DDPG which is trained from scratch, and the policy obtained from the method described in Section 3.3. As Fig. 4 shows, eventually, our proposed method achieves a considerable performance boost from the original imitation learning's learned policy while the pure DDPG never performs well and does not show any improving trend.

A relevant simulation video can be found at http://youtu.be/yjmM70alCSQ.

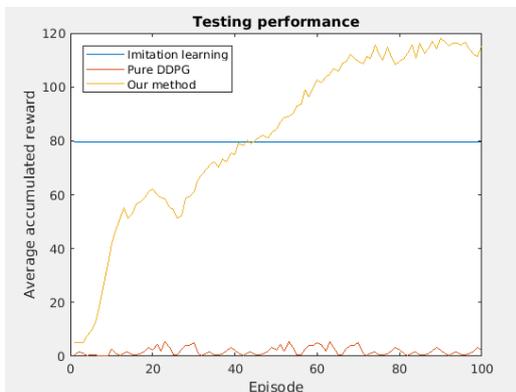

Fig. 4 Comparison of the testing performance

## 5. CONCLUSION

In this paper, we have presented a method which combines the advantages of both imitation learning and reinforcement learning with the utilization of pre-trained weights from imitation learning and modified DDPG algorithm. Through training in diverse environments with various weather and lighting conditions, we have eventually got a robust driving policy with boosted performance comparing to the generated policy through imitation learning. Since humans are relatively good at driving cars based on the input images, for the autonomous driving task, leveraging human's guidance as compensation of pure reinforcement learning has shown good performance in this task and might be a promising direction of autonomous driving.

## ACKNOWLEDGMENT

This research has been supported in part by the KUSTAR-KAIST Institute, KAIST, Korea and the ICT R&D program of MSIP/IITP [2016-0-00563, Research on Adaptive Machine Learning Technology Development for Intelligent Autonomous Digital Companion].